\documentclass[conference]{IEEEtran}
\usepackage{fancyhdr}
\usepackage{stfloats}
\usepackage{xcolor}
\usepackage{colortbl}
\usepackage[most]{tcolorbox}

\usepackage{url}
\usepackage{setspace}
\usepackage{placeins}
\usepackage{cite}

\usepackage{algorithm}
\usepackage[noend]{algpseudocode}
\algnewcommand{\LineComment}[1]{\State \emph{\textcolor{blue}{\(\triangleright\) #1}}}
\algrenewcommand\algorithmicindent{1em}%

\usepackage{graphicx}
\usepackage{caption}
\usepackage{subcaption}
\usepackage{tabularx}
\usepackage{enumitem}

\usepackage{color, soul}
\soulregister\emph7
\soulregister\cite7
\soulregister\ref7
\soulregister\pageref7

\definecolor{Gray}{gray}{0.85}
\definecolor{LightCyan}{rgb}{0.88,1,1}

\begin{document}

\newfloat{lstfloat}{htbp}{lop}
\floatname{lstfloat}{Listing}
\def\lstfloatautorefname{Listing}

\setlength{\abovedisplayskip}{-1pt}
\setlength{\belowdisplayskip}{-1pt}

\setlength\floatsep{0.3\baselineskip plus 1pt minus 20pt}
\setlength\textfloatsep{0.3\baselineskip plus 1pt minus 20pt}
\setlength\intextsep{0.3\baselineskip plus 1pt minus 20pt}
\newcolumntype{R}{>{\centering\arraybackslash}m{3.5cm}}
\newcolumntype{L}{>{\centering\arraybackslash}m{1.5cm}}
\newcolumntype{M}{>{\centering\arraybackslash}m{3.5cm}}

\title{Pushing the Envelope of LLM Inference with Ultra-Low-Bit Quantized Models}
\author{\IEEEauthorblockN{Evangelos Georganas,
Dhiraj Kalamkar, Alexander Heinecke, Pradeep Dubey}
\IEEEauthorblockA{Intel Corporation}}
\maketitle

\begin{abstract}
The advent of ultra-low-bit LLM models, approaching the perplexity and task accuracy of their full precision counterparts, is ushering in a new era of LLM inference. While these advances promise models that are cost-effective regarding latency, memory, throughput, and energy consumption, the efficiency of runtimes for deploying ultra-low-bit models remains under-explored. In this work, we take a bottom-up approach: we first implement 2-bit microkernels for modern CPUs, achieving close-to-roofline performance. We integrate these microkernels into LLM inference pipelines and present end-to-end results with 2-bit models, outperforming the state-of-the-art (SOTA) bitnet.cpp runtime by 2.2$\times$, and deliver up to 7$\times$ speedup compared to 16-bit inference. We extend this work to Intel Xe2 GPUs where we implement mixed-precision, 2-bit kernels, and show their performance to be close-to-optimal. We integrated the GPU kernels in the vLLM framework and evaluated end-to-end inference for a range of models and Xe2 GPUs. We obtain up to 6.7$\times$ speedup compared to the 16-bit pipeline, pushing the envelope of LLM inference.
\end{abstract}

\section{Introduction}
\label{sec:introduction}
Recent advances in Quantization-Aware Training (QAT) of Large Language Models (LLM) have produced ultra-low-bit (1-bit, 1.58-bit and 2-bit) quantized models that come within 5--10\% of the perplexity and end-task performance of their full-precision counterparts~\cite{1bit, liu2025paretoq, prismml2026bonsai}. In particular, recent work~\cite{liu2025paretoq, prismml2026bonsai} shows that 2-bit and ternary models reside on the Pareto frontier in terms of the accuracy-model size trade-off. When considering the metric \emph{intelligence density} as intelligence per unit of storage, the ternary Bonsai models exhibit almost a 10$\times$ increase over the 16-bit counterparts~\cite{prismml2026bonsai}. Such compact LLMs have the potential to be cost-effective in latency, memory, throughput, and energy consumption in the era of agentic AI systems~\cite{belcak2025small}. For example, the ternary Bonsai 27B model is within $\sim$5\% accuracy of the 16-bit Qwen3.6-27B, offering multi-step reasoning, structured tool calls, vision tasks, and agentic loops capablities with $\sim$9$\times$ smaller footprint compared to the 16-bit Qwen3.6-27B~\cite{prismml2026bonsai}. Despite these advancements in the quantization of LLMs, the inference runtimes capable of serving ``ultra-low-bit'' quantized models are limited. The state-of-the-art (SOTA) inference runtime bitnet.cpp is designed for ternary LLMs, with a relatively narrow range of applicable model architectures~\cite{bitnet}. Bitnet.cpp improves upon alternatives (e.g.\ llama.cpp~\cite{llamacpp}) being up to 6$\times$ faster by introducing a specialized mixed-precision matrix multiplication library to facilitate sub-2-bit models. However, its efficiency is not evaluated in terms of optimality or roofline performance. Our preliminary performance analysis showed that 2-bit inference with bitnet.cpp was even \emph{slower} than SOTA 4-bit inference on CPUs (e.g.\ see in Figures~\ref{fig:e2e_arl},~\ref{fig:e2e_arlh},~\ref{fig:e2e_lnl} orange and magenta bars), highlighting that bitnet.cpp is far from optimal.

In this work we adopt a bottom-up approach to address the performance shortcomings of current ultra-low-bit LLM inference. Client-side/edge inference typically involves single- or small-batch use cases, which in turn result in memory-bound matrix multiplication invocations during the execution. By exploiting the reduced bit-width of the model's weight tensors during inference we can accelerate the corresponding matrix multiplications (compared to the case with 16-bit weight tensors), since the volume of data read on the critical path is proportional to the bit-width of the weight datatype. First, we design and implement 2-bit mixed-precision matrix multiplication (GEMM) routines that up-convert the low-precision weight matrices (2-bit) to 8-bit integers, and perform the fused-multiply-add (FMA) operations using hardware-accelerated instructions on modern CPUs. We introduce novel weight layouts that facilitate the up-convert process, and within the GEMM microkernels we deploy instruction sequences that maximize the throughput of the ``up-convert and compute'' steps. We also construct performance models for the introduced kernels to assess their efficacy and show that they achieve close-to-peak roofline performance. We compose these microkernels into optimized multi-threaded GEMM routines that leverage dynamic task scheduling to fully exploit the heterogeneous architectures with \emph{performance} and \emph{efficiency} cores. Then, we integrate these ultra-low-bit GEMM routines in a SOTA inference framework, namely PyTorch TPP~\cite{georganas2021tensor}, and showcase end-to-end inference results using 2-bit models. Our optimized 2-bit inference is up to 7$\times$ faster than the 16-bit inference, and outperforms bitnet.cpp by up to 2.2$\times$.

We extend this work to Intel Xe2 GPUs where we design and implement mixed precision, 2-bit GEMM kernels. While contemporary Intel Xe2 GPUs support natively int2$\times$int8$\rightarrow$int32 computations using hardware acceleration, the SOTA ultra-low-bit inference interfaces with high-precision (e.g.\ BF16) input/output activations, i.e.\ int2$\times$BF16$\rightarrow$BF16 GEMM. Quantizing the input activations from BF16 to int8 \emph{before} the GEMM operation, and dequantizing the int32 GEMM output activations to BF16 via separate microkernels out of L1/L2 cache adds minimal overhead on CPUs. However, on GPUs such a decoupled execution of activation-quantization $\rightarrow$ low-precision GEMM microkernel $\rightarrow$ output-dequantization incurs multiple kernel launches and passes over the data. Therefore, we developed mixed precision int2$\times$BF16$\rightarrow$BF16 GEMM Xe2 kernels that fuse the input/output activation quantization/dequantization within the GEMM kernel to minimize the corresponding overheads. We integrated our Xe2 kernels in the vLLM~\cite{vllm} framework as a quantization plugin and evaluated end-to-end LLM inference results for a range of LLM models and Intel Xe2 GPUs. Depending on the model and platform, we see up to $8\times$ reduction in GEMM time compared to the BF16 case, and we get up to 6.7$\times$ speedup in end-to-end inference compared to the BF16 execution.

The contributions of this work are:
\begin{itemize}[labelindent=0em,labelsep=0.2cm,leftmargin=*,noitemsep,topsep=0pt]
\item We designed and implemented 2-bit GEMM kernels tailored for CPUs with AVX2 ISA, that leverage novel blocked-interleaved weight-tensor layouts. We devised performance models that characterize the effectiveness and limitations of these kernels. We composed these microkernels into multi-threaded GEMM routines targeting hybrid-core architectures. We benchmarked these routines on a set of matrices and platforms, achieving close-to-peak memory bandwidth.
\item We designed and implemented novel mixed precision GEMM kernels targeting Intel Xe2 GPUs. These kernels fuse the quantization/dequantization of the input/output activations and leverage the hardware-accelerated int2$\times$int8 Dot Product Accumulate Systolic (DPAS) instructions. We benchmarked the mixed precision GEMM kernels and show they achieve bandwidth close to the platform's limit, accelerating memory-bound GEMMs proportionally to the bit-width of the weight's datatypes. We achieve compute throughput close-to-ideal int2$\times$int8 DPAS throughput, showing the efficacy of fused quantization/dequantization, and these Xe2 kernels can be used to accelerate compute-bound cases in LLM inference (e.g.\ prefill, large batch size).
\item We integrated these ultra-low-bit GEMM kernels in SOTA end-to-end inference pipelines (PyTorch-TPP for CPU, vLLM for GPU), assessed their performance on a range of models and platforms, and observed speedups up to 7$\times$ over the 16-bit inference. We benchmarked our runtime against bitnet.cpp and show speedups up to 2.2$\times$.
\item We compare the CPU performance with the GPU 2-bit inference within bitnet.cpp, which includes CUDA-optimized 2-bit kernels. The performance on contemporary x86 client CPUs is within 3.7$\times$--4.9$\times$ of the RTX PRO 6000 Blackwell Server Edition GPU (henceforth called B6000), which has 18$\times$--24$\times$ more bandwidth than the CPU platforms. On a discrete Intel Arc B70 Xe2 GPU we achieve a speedup of 1.34$\times$ over the 2-bit inference on the NVIDIA B6000 GPU for a same-sized 2 billion parameter model, despite NVIDIA B6000 having 3$\times$ more bandwidth than Intel B70.
\end{itemize}

\section{Ultra-low-bit GEMM microkernels}
\label{sec:microkernels}
\subsection{Overview and background}
In this section we detail the ultra-low-bit GEMM microkernels, that take as input a 2-bit (int2) weight tensor, multiply it with the input activation tensor and produce the output activations. We refer to this technique as ``up-convert and compute''. We focus on modern CPUs with AVX2 instructions that can be found on contemporary AI-PC devices, and also on modern Intel GPUs with Xe2 cores that natively support int2$\times$int8 operations. We target the use-cases where the GEMMs exhibit algorithmically low arithmetic intensity, thus are bandwidth-bound (e.g.\ single/low-batch inference). For the CPU case, we implemented all the GEMM microkernels using the libxsmm library~\cite{heinecke2016libxsmm, georganas2021tensor}. Our multi-threaded GEMMs for CPUs use these JIT-ed microkernels as building blocks, and for parallelization we use PARLOOPER and its hybrid scheduling to fully utilize all performance and efficiency cores~\cite{georganas2024harnessing}. For the Xe2 GPUs, we implement the corresponding low-bit microkernels using the XeTLA framework: a collection of SYCL/ESIMD templates that enable high-performance computations on Intel Xe GPU architectures~\cite{xetla}.

\subsection{Interleaved Tensor Layout for 2-bit Weights on CPU}
\label{subsec:layouts}
\begin{figure*}
\centering
\includegraphics[width=1.0\textwidth]{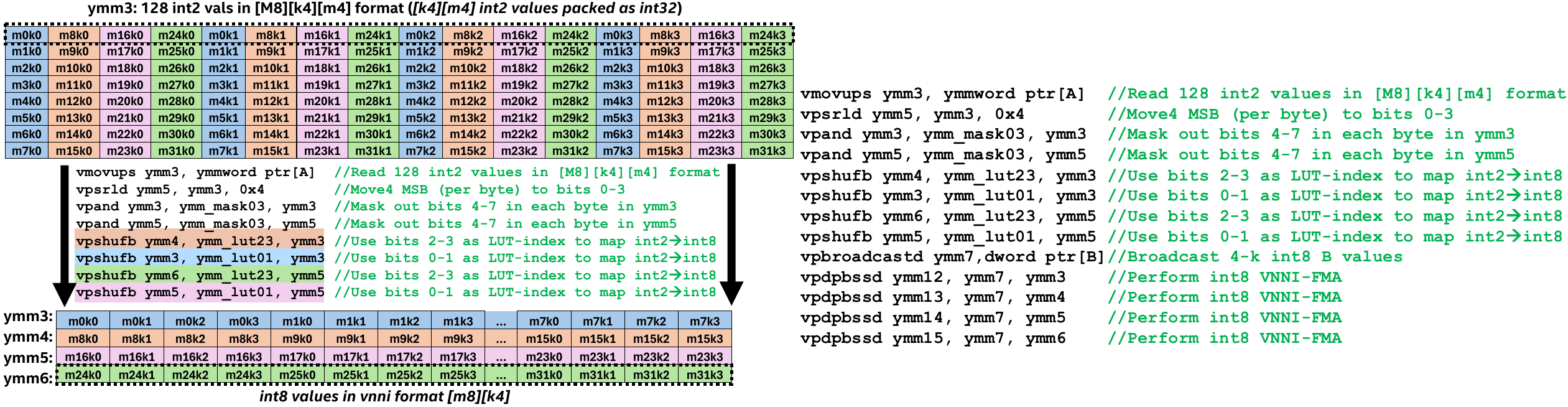}
\caption{\textbf{Left:} Unpacking the int2 VNNI4-interleaved format to int8 VNNI4. We pack a $[k4][m4]$ 2-bit subtensor in a 32-bit value (see dotted sub-tensor in top) and with a full vector load of 256-bits we can read 128 2-bit entries, which effectively form an $[M8][k4][m4]$ tensor. With 1 logical shift + 2 logical AND + 4 byte-shuffles we get as output 4 256-bit vectors, each holding an $[m8][k4]$ int8 subtensor which is in VNNI4 layout. \textbf{Right:} AVX2 GEMM microkernel with int2 weights (matrix $A^{M\times K}$), int8 activations (matrix $B^{N\times K}$) and VNNI-int8 FMAs ($M=32$, $N=1$, $K=4$).}
\label{fig:unpack}
\end{figure*}
Naively up-converting the int2 weights to FP32 values (e.g.\ via look-up tables or permutes) and then multiplying with the input activations requires an excessive number of operations, which diminishes the benefits of reading the low-bit data type. CPUs on modern client platforms have a limited number of cores (e.g.\ 10-20), yet offer bandwidth on the order of 100 GB/s. The available bandwidth per core is quite high, thus simple FP32 FMA throughput is insufficient to saturate the per-core memory bandwidth considering that we read $8\times$ less data than the BF16-equivalent GEMMs while performing the same number of compute operations.

To address this, we quantize the input activations to int8, enabling the int8 VNNI FMAs available on modern AVX2 CPUs that offer $4\times$ the throughput of FP32 FMAs. The input activation \emph{quantization}, along with the required output activation \emph{dequantization}, can be implemented as separate operations before/after the quantized GEMM operations with minimal overhead on the CPUs. More specifically, for memory-bound matrix-vector multiplication scenarios the input and output activations are merely vectors with small size compared to the weight tensors, and as such their quantization and dequantization incur negligible overhead (especially considering that dequantization happens out of L1/L2 cache). The int8 VNNI FMA requires the input vector operands to pack together four int8 values from the contraction dimension of the GEMM (which we denote as $K$ throughout the paper), and this is the so-called VNNI4 layout. Naively packing four 2-bit values in the innermost dimension of the weight tensor to form a VNNI4 layout requires during runtime an excessive number of shifts and shuffles to (i) unpack the int2 values to int8, and (ii) to re-pack the int8 values back to VNNI4 layout (with a two-level shuffle network) for the upcoming int8-VNNI FMAs.

To alleviate this unpacking bottleneck, we introduce a new tensor layout called VNNI4-interleaved, illustrated in Figure~\ref{fig:unpack}-Left. In this layout, a logical $M\times K = m32\times k4$ tensor is transformed into an interleaved 3D tensor by folding in the innermost dimension entries from the logical $M$ dimension that are iso-modulo 8. The new tensor layout reflects a 3D tensor $M8\times k4\times m4$ where the innermost folded $m4$ dimension corresponds to the originally-indexed $m32$ entries with the same modulo 8 index. In this way, we pack a $[k4][m4]$ 2-bit subtensor in a 32-bit value (see dotted sub-tensor in top of Figure~\ref{fig:unpack}-Left), and with a 256-bit vector-load we can read 128 2-bit entries, which effectively form an $[M8][k4][m4]$ tensor.

\subsection{2-bit GEMM microkernel for CPU}
\label{subsec:2bit}
Using our new 2-bit layout we can simply perform table-lookups that utilize the 2-bit entries as indices and translate them to int8 values that effectively form within a 256-bit vector an $[m8][k4]$ int8 subtensor in VNNI4 format. This int8 vector can be readily used as input to int8 VNNI FMA. In the example of unpack instruction sequence (between the arrows in Figure~\ref{fig:unpack}-Left), with the byte-shuffle highlighted in blue \texttt{vpshufb ymm3, ymm\_lut01, ymm3} we consider the 32 2-bit indices highlighted in blue from the original 2-bit packed vector. The \texttt{ymm\_lut01} considers only the bits 0-1 of each byte in the vector and maps these indices to the corresponding int8 values. In an analogous way, with the byte-shuffle highlighted in orange \texttt{vpshufb ymm4, ymm\_lut23, ymm3} we consider the 32 2-bit indices with orange color from the original 2-bit packed vector. The \texttt{ymm\_lut23} considers only bits 2-3 of each byte in the vector and maps these indices to the proper int8 values. The shift instruction \texttt{vpsrld ymm5, ymm3, 0x4} shifts the 4 higher bits 4-7 within each byte to the 4 lower positions such that with 2 more shuffles (highlighted with green and magenta) we can unpack the corresponding int2 values to int8 values. The 2 logical AND vector operations (\texttt{vpand}) in the unpack instruction sequence merely mask out the highest 4 bits within each byte to ensure that the byte-shuffles have the desired effect. After this instruction sequence with 1 logical shift + 2 logical AND + 4 byte-shuffles we get as output 4 256-bit vectors, each holding an $[m8][k4]$ int8 subtensor which is in VNNI4 layout.

The aforementioned byte permutes/shuffles are fast on modern CPU architectures since the shuffled values do not cross 128-bit lanes within a vector. Also, the input int2 tensor is packed in this interleaved format off-line since during inference the weight tensors are ``frozen'' and we can freely pack/reformat them ahead of time. Even though we described how we reformat a logical $32\times 4$ 2-bit vector to a $[M8][k4][m4]$ layout, for the entire weight tensor we adopt a blocked layout. The input weight tensor $W$ is conceptually a 2D matrix $W^{M\times K}$. We follow the approach of previous work~\cite{georganas2020harnessing} and we block the dimensions $M$ and $K$ by factors $b_m=32$ and $b_k$ respectively. Such a blocked layout exposes better locality and avoids large, strided sub-tensor accesses which are known to cause TLB misses and cache conflict misses~\cite{georganas2020harnessing}. By using these blocking factors $b_m=32$ and $b_k$, the logical $[K][M]$ weight tensor is blocked and interleaved in a $[M/32][K/b_k][b_k/4][M8][k4][m4]$ layout. We leverage the BRGEMM TPP in order to perform the tensor contraction over dimensions $K/b_k$ and $b_k$, which constitute the inner-product dimension of the original 2D matrix~\cite{georganas2020harnessing}.

With such a weight format, we can readily design a mixed precision GEMM microkernel that uses 2-bit weights, 8-bit activations and produces 32-bit outputs (see Figure~\ref{fig:unpack}-Right). First, with a full 256-bit vector load we load 128 int2 values in $[M8][k4][m4]$ layout in vector register \texttt{ymm3}. Then with the up-convert sequence we unpack these 128 int2 values to 128 int8 values occupying 4 vector registers \texttt{ymm3} - \texttt{ymm6}. With a 32-bit broadcast \texttt{vpbroadcastd} we broadcast in \texttt{ymm7} four 8-bit values from the input activations that correspond to 4 elements from the logical $K$ dimension. Last, with four VNNI-int8 FMAs (\texttt{vpdpbssd}) we perform the corresponding multiply-add operations among vectors \texttt{ymm3} - \texttt{ymm6} and vector \texttt{ymm7} and we accumulate the partial products in the output 32-bit accumulators \texttt{ymm12} - \texttt{ymm15}. This microkernel effectively multiplies a logical $32\times 4$ $A$ subtensor with a $4 \times 1$ $B$ vector and produces an output $32\times 1$ with 32-bit integer precision. Using this microkernel as a building block, we can compose microkernels with larger $M$, $K$ and $N$~\cite{heinecke2016libxsmm, georganas2020harnessing}.

\subsection{2-bit GEMM kernels for Intel Xe2 GPU}
\label{subsec:gpu_kernels}
\begin{figure*}
\centering
\includegraphics[width=1.0\textwidth]{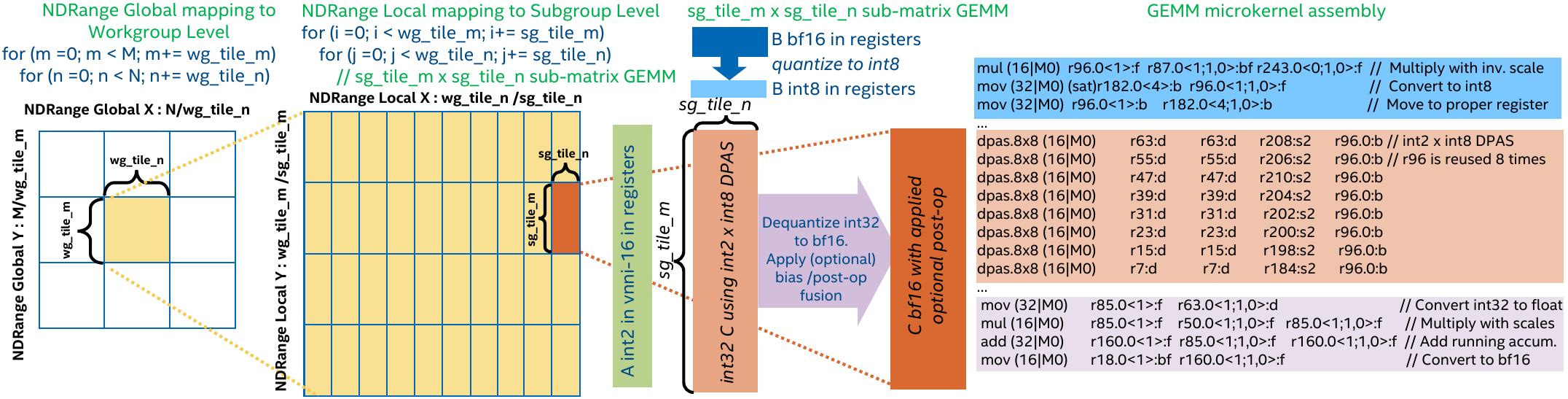}
\caption{Xe2 GPU GEMM kernel with int2 weights ($A$) and BF16 activations ($B$ and $C$). We split the output matrix $C^{M\times N}$ into tiles with size $wg\_tile\_m\times wg\_tile\_n$ and each workgroup will calculate a sub-matrix $wg\_tile\_m\times wg\_tile\_n$ (see yellow box at Left). Subsequently, this sub-matrix will be further divided into multiple tiles, with dimensions $sg\_tile\_m\times sg\_tile\_n$ (see dark-orange rectangular tile). These tiles will then be assigned to subgroups. Finally, the corresponding subgroup GEMM microkernel and the involved tile operations will be mapped to the actual Xe2 instructions, such as 2D-load and DPAS instructions (see GEMM microkernel at Right). The quantization of $B$ is fused in the GEMM: while the loaded $B$ sub-matrix is in BF16 (dark-blue box), we quantize the entries to int8 using in-register operations (light-blue assembly box). With rectangular shapes for the $sg\_tile\_m\times sg\_tile\_n$ tiles we re-use the quantized $B$ and amortize the corresponding overhead.}
\label{fig:gpu_kernel}
\end{figure*}
Intel Xe2 GPUs support natively int2$\times$int8$\rightarrow$int32 GEMM computations using hardware-accelerated Dot Product Accumulate Systolic (DPAS) instructions. Therefore, for the Xe2 int2 kernels we use these instructions without the need to up-convert the weights to int8 before performing the necessary multiply-add operations (cf.\ the CPU up-convert \& compute methodology). Nevertheless, the SOTA ultra-low-bit inference (e.g.\ 2-bit weights) requires high-precision (e.g.\ BF16) input/output activations, i.e.\ the involved GEMM precision is int2$\times$BF16$\rightarrow$BF16. For the CPU case, quantizing the input activations from BF16 to int8 \emph{before} the int2$\times$int8$\rightarrow$int32 GEMM operation, and dequantizing the int32 GEMM output to BF16 output activations via separate microkernels out of L1/L2 cache adds minimal overhead. More specifically, for memory-bound matrix-vector multiplication the input and output activations are merely vectors with small size compared to the weight tensors, therefore their quantization and dequantization incur negligible overhead since these operations run out of L1/L2 cache. However, on GPUs such a decoupled execution (i.e.\ multiple kernels) of activation-quantization $\rightarrow$ low-precision GEMM microkernel $\rightarrow$ output-dequantization incurs multiple kernel launches and passes over the data.

To mitigate the shortcomings of multiple kernel launches, we developed mixed precision int2\ $\times$ BF16$\rightarrow$BF16 GEMM Xe2 kernels that fuse the input/output activation quantization/dequantization within the GEMM kernel (see Figure~\ref{fig:gpu_kernel}). Using conventional SYCL terminology~\cite{sycl}, we split the output matrix $C^{M\times N}$ into tiles with size $wg\_tile\_m\times wg\_tile\_n$ and each workgroup will calculate a sub-matrix $wg\_tile\_m\times wg\_tile\_n$, represented by the yellow box in output $C$ (see Left part of Figure~\ref{fig:gpu_kernel}). This sub-matrix is subsequently divided into multiple tiles, with dimensions $sg\_tile\_m\times sg\_tile\_n$. These tiles are assigned to subgroups. Finally, the corresponding subgroup GEMM microkernel and the involved tile operations are mapped to the actual Xe2 hardware instructions, such as 2D-load and DPAS instructions. The int2$\times$int8$\rightarrow$int32 DPAS instructions for Xe2 require the int2 operand to be formatted in VNNI16 layout: pack together 16 int2 values from the logical $K$ dimension of the GEMM. Since we are performing inference we merely pre-format the int2 weights to such VNNI16 layout without any overhead during runtime.

Figure~\ref{fig:gpu_kernel} also illustrates the fusion of the input activation (matrix $B$) quantization and the output dequantization. When loading the matrix $B$ from memory in registers, the $B$ precision is BF16 (see dark blue box). In order to quantize the BF16 $B$ entries given some scales $B\_scales$ (see blue box in the GEMM assembly), we first multiply the entries with the inverse of the scales, then we convert the result to int8 and these int8 values (light-blue box) can be readily used in the DPAS instructions (see peach-colored box in assembly). By carefully selecting the shapes of the $sg\_tile\_m\times sg\_tile\_n$ tiles we can re-use the quantized $B$ entries and amortize the corresponding overhead (for the rectangular $sg\_tile\_m\times sg\_tile\_n$ tiles of Figure~\ref{fig:gpu_kernel} we re-use each quantized $B$ register 8 times). We iterate the same process across all the tiles in the $K$ dimension. Once the full accumulation has been completed, we need to dequantize the int32 output registers forming a logical $sg\_tile\_m\times sg\_tile\_n$ sub-tensor. First we convert the int32 values to float, then we multiply these float values with the \emph{dequantization} scales and finally we convert the result to BF16 that can be stored in the output $C$ in memory e.g.\ via 2D-store Xe2 operations (see purple box in microkernel assembly). Also any additional operations (e.g.\ bias addition, post-GEMM activation functions) can be fused at this point where the $C$ entries are still in-registers before storing them to memory. Our Xe2 GEMM implementation also supports $K$-splitting algorithms that extract parallelism across the $K$ dimension for smaller GEMM shapes~\cite{xetla}.

With respect to the calculation of the $B\_scales$, we have to compute the absolute maximum per quantization granularity (e.g.\ the absolute max of $K$ $B$ entries or some other user-defined granularity). For this abs-max operation we implemented an optimized SYCL reduction kernel that can be either performed before the GEMM as a separate kernel or it can be fused at the work-group level, and the calculated abs-max values are stored in fast Shared Local Memory (SLM) available on Xe2. Fusing the abs-max $B$ reduction in the GEMM kernel avoids one kernel-launch overhead at the expense of redundant computations (multiple work-groups might perform the same reduction in case they share the same slabs of $B$). We expose this option of fusing/non-fusing $B$ scale calculation as a tuning knob. The $C$ \emph{dequantization} scales mentioned earlier are just a product of the $B\_scales$ and the $A\_scales$ that are part of the input int2 $A$ weight matrix.

In order to extract the maximum performance of our Xe2 kernels for a variety of GEMM shapes, we developed an auto-tuning framework within XeTLA that searches for optimal parameter configurations, e.g.\ fusing/non-fusing $B$ scale calculation, optimal $wg\_tile\_m$, $wg\_tile\_n$, $sg\_tile\_m$, $sg\_tile\_n$ tile sizes, using/not-using $K$-splitting algorithms etc.

\paragraph*{Alternative implementation for 2-bit Xe2 kernel}
\begin{figure}[t]
\begin{tcolorbox}[colback=black!3,boxrule=0.4pt,left=2pt,right=2pt,top=2pt,bottom=2pt]
\scriptsize
\setlength{\tabcolsep}{0pt}
\renewcommand{\arraystretch}{0.96}
\def\sect#1{\multicolumn{1}{@{}l@{}}{\texttt{;}\,\textrm{\emph{#1}}}}
\newcommand{\lt}{\textless}
\newcommand{\gt}{\textgreater}
\begin{tabular}{@{}>{\ttfamily}l@{}}
\sect{r10: packed 2-bit codes, r27: fp16 scale of this $K$-group} \\[1pt]
\sect{even bit (code LSB) $\rightarrow$ predicate} \\
and (16\textbar M0) (eq)f0.1 null\lt2\gt:w r10.0\lt2;1,0\gt:w 1:w \\[1pt]
\sect{odd bit (code MSB) $\rightarrow$ the fp16 sign position} \\
shl (16\textbar M0) r28.0\lt2\gt:w r10.0\lt2;1,0\gt:w 14:w \\
mov (16\textbar M0) r29.0\lt1\gt:w r28.0\lt2;1,0\gt:w \\
mov (16\textbar M0) r30.0\lt1\gt:hf r29.0\lt1;1,0\gt:hf \\[1pt]
\sect{XOR sign into the scale's fp16 bit pattern} \\
bfn.(s0\textasciicircum s1\&s2) (16\textbar M0) r30.0\lt1\gt:uw r27.0\lt1;0\gt:uw \\
\hspace*{1.2em}r30.0\lt1;0\gt:uw 0x8000:uw \\[1pt]
\sect{force zero where the even bit is clear} \\
($\sim$f0.1) sel (16\textbar M0) r31.0\lt2\gt:w r30.0\lt1;1,0\gt:w 0:w \\
\end{tabular}
\end{tcolorbox}
\caption{Alternative Xe2 sequence turning 2-bit ternary values to fp16 weights by mixing 2-bit codes with fp16 weight scales.}
\label{fig:xe2int2f16asm}
\end{figure}

Even though we focused on the optimized fused Xe2 implementation that uses int2$\times$int8$\rightarrow$int32 computations and should be able to achieve close to the corresponding DPAS peak, our XeTLA implementation supports alternative kernels for the int2 GEMM. In Figure~\ref{fig:xe2int2f16asm} we show an implementation where the input 2-bit (ternary) weights are mingled with the corresponding 16-bit weight scales to form 16-bit weight values which are readily fed to a 16-bit$\times$16-bit$\rightarrow$32-bit DPAS. In such a kernel, there is no need to quantize the input activations or dequantize the output after the DPAS instructions. In the Xe2 assembly depicted at Figure~\ref{fig:xe2int2f16asm}, the 2 bits of a ternary weight are separated: the even bit (LSB) becomes a predicate \texttt{f0} (dictating if this is a zero or a non-zero weight), and the odd bit (MSB) dictates the sign of the corresponding weight. The odd bit is shifted into bit 15 so that \texttt{bfn} conditionally sets the sign bit of the 16-bit weight group scale (originally in register \texttt{r27}). The fp16 weight is therefore reconstructed by an XOR, and the predicated \texttt{sel} by \texttt{f0} yields a vector with the proper values $\{-weight\_scale, 0, +weight\_scale\}$ that is consumed by an fp16$\times$fp16$\rightarrow$fp32 DPAS. By construction, this Xe2 kernel leaves the activations unquantized and yields fp32 output values, thus no dequantization is required. Even though this kernel with fp16$\times$fp16$\rightarrow$fp32 DPAS compute has 2$\times$ lower compute peak compared to the kernel with the int2$\times$int8$\rightarrow$int32 DPAS instructions, in memory bound scenarios it should also be able to saturate the available bandwidth.

\subsection{Performance modeling \& limitations of the CPU kernels}
\label{subsec:perf_model}

The low-precision GEMMs on CPU require additional work to up-convert the low-precision operand to int8 since there is no native hardware support (unlike Xe2 GPUs). Also, the aggregate int8 FMA throughput on client CPUs with a \emph{limited} number of cores poses a challenge in achieving peak bandwidth. We devise roofline models for the 2-bit CPU microkernels to assess their efficacy and limitations. We aim to determine under which conditions the CPU microkernels can saturate the bandwidth and effectively accelerate matrix-vector multiplication during inference. We follow the reasoning/bottleneck-style analysis of the original roofline model~\cite{roofline}. Let $\beta$ denote the number of bytes/cycle each core in a multi-core CPU can read when all cores access memory simultaneously. Let $\gamma$ represent the number of cycles to \emph{compute} an output 256-bit vector in the GEMM without considering the data load instructions. Assuming $B$ bytes must be read from memory to compute an output 256-bit vector, the number of cycles $T$ to calculate an output vector is:
\begin{equation}
    T = \max(B/\beta, \gamma)
\end{equation}
To operate in a bandwidth-bound regime, since we are reading the weight tensor from memory and the matrix-vector multiplication has no weight re-use, we need the first term in the max operator above to be the dominant one, i.e.:
\begin{equation}
    \gamma \le B/\beta \Rightarrow \beta \le B/\gamma
    \label{eqn:ineq}
\end{equation}
We can determine the effective bandwidth per core $e_{bw}$ to be bounded by the minimum of the terms $\beta$ and $B/\gamma$ (the latter term representing the throughput of the ``up-convert \& compute'' sequence). In short we can write:
\begin{equation}
    e_{bw} = \min(\beta, B/\gamma)
    \label{eqn:ebw}
\end{equation}
\begin{figure}[t]
\centering
\includegraphics[width=1.0\columnwidth]{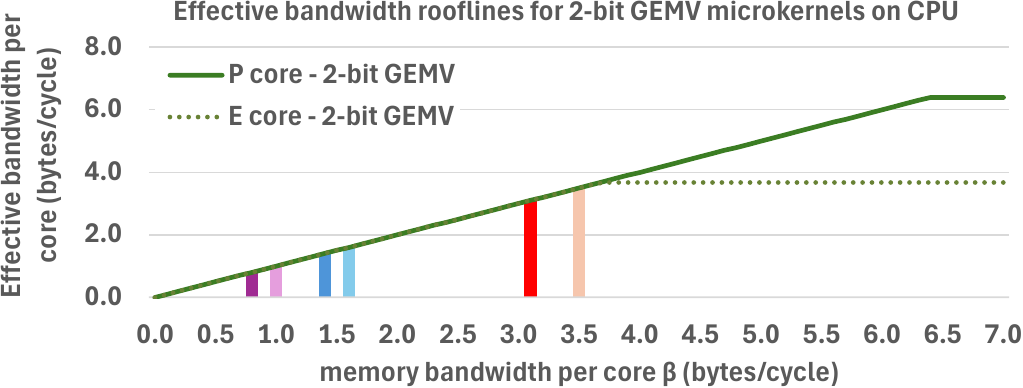}
\caption{Effective bandwidth rooflines for 2-bit GEMV microkernels considering $P$ and $E$ cores. The y-axis shows the effective bandwidth per core in bytes/cycle and the x-axis shows the available memory bandwidth per core $\beta$ in bytes/cycle. The vertical lines depict the $\beta$ values for the different CPU platforms used in our experiments (see Section~\ref{sec:results}).}
\label{fig:roofline}
\end{figure}
Since our CPU platforms have performance $P$ and efficiency $E$ cores with different compute capabilities, we consider two different $\gamma$ factors, namely $\gamma_P$ and $\gamma_E$, and consequently two different effective bandwidth terms $e_{bw}^P = \min(\beta, B/\gamma_P)$ and $e_{bw}^E = \min(\beta, B/\gamma_E)$. For the 2-bit GEMM microkernel, the $\gamma$ factor entails the cycles to perform (1 logical shift + 2 logical AND + 4 byte-shuffles + 4 int8 FMAs) / 4. We are interested in the throughput of these instruction sequences given their dependencies, and we determined these $\gamma$ values empirically for both $P$ and $E$ cores on contemporary x86 client CPUs. Through micro-benchmarks we find that $\gamma_P =1.25$ and $\gamma_E= 2.175$. By using these values in Equation~\ref{eqn:ebw} and with $B=8$ (we read 8 bytes to compute an 256-bit output vector), we get: $e^P_{bw} = \min(\beta, 6.4)$ and $e^E_{bw} = \min(\beta, 3.68)$. Figure~\ref{fig:roofline} shows the rooflines for the effective bandwidth per core in bytes/cycle (y-axis) while the x-axis shows the available memory bandwidth per core $\beta$ in bytes/cycle. The solid green line depicts $e^P_{bw}$ for the 2-bit microkernel while the dashed green line corresponds to the $e^E_{bw}$. In the next section we use this model to analyze the efficacy of our microkernels on CPUs with $\beta$ values annotated with vertical bars in Figure~\ref{fig:roofline}.

This roofline also highlights the limitations on CPUs: when the $\gamma$ increases (e.g.\ reduced FMA throughput, multiple activation vectors requiring
more FMAs per loaded weight-vector) or the bandwidth per core $\beta$ increases (e.g.\ increased total bandwidth, or reduced number of cores), then it becomes more challenging to achieve effective bandwidth $e_{bw}$ close to $\beta$.

\section{Results}
\label{sec:results}
\subsection{Experimental Platforms}
We use three contemporary x86 CPUs with varying number of performance/efficiency cores, and memory bandwidth:
\begin{itemize}[labelindent=0em,labelsep=0.2cm,leftmargin=*,noitemsep,topsep=0pt]
\item Intel Core Ultra 9 285K CPU (referred to as ARL) with 24 cores (8 P and 16 E cores). It has overclocked memory DDR5@7200 MT/s, measured read bandwidth of 98 GB/s and Maximum Turbo Power of 250~W.
\item Intel Core Ultra 7 255H CPU (referred to as ARLH) with 14 cores (6 P and 8 E cores). It has memory DDR5@5600 MT/s with measured read bandwidth of 75 GB/s and Maximum Turbo Power of 115~W.
\item Intel Core Ultra 7 258V CPU (referred to as LNL) with 8 cores (4 P and 4 E cores). It has 32 GB of DDR5@8533 MT/s memory with measured read bandwidth of 97 GB/s and Maximum Turbo Power of 37~W.
\end{itemize}

For our GPU evaluation we use two Xe2 GPU systems:
\begin{itemize}[labelindent=0em,labelsep=0.2cm,leftmargin=*,noitemsep,topsep=0pt]
\item Intel Arc 140V, which is the integrated GPU in the aforementioned Lunar Lake (LNL) platform. It consists of 8 Xe2 GPU cores and it has access to the same memory/bandwidth as the LNL CPU (32 GB of DDR5@8533 MT/s with measured read bandwidth of 97 GB/s).
\item Intel Battlemage Arc B70 (BMG) discrete GPU. It has 32 Xe2 cores, 32 GB of GDDR6 memory, delivering up to 608 GB/s. The int8 peak is 367 TOPS (Trillions Operations per Second) and its Total Board Power (TBP) is 230~W.
\end{itemize}

For all the presented results (microbenchmarks and end-to-end inference) we measure little to no variation (typically less than 3\%). Thus we merely run 10-20 iterations (depending on the experiment) and we report the average time/performance.

\subsection{GEMV microbenchmarks on CPU}
\label{subsec:micros}
\begin{figure}[t]
\centering
\includegraphics[width=1.0\columnwidth]{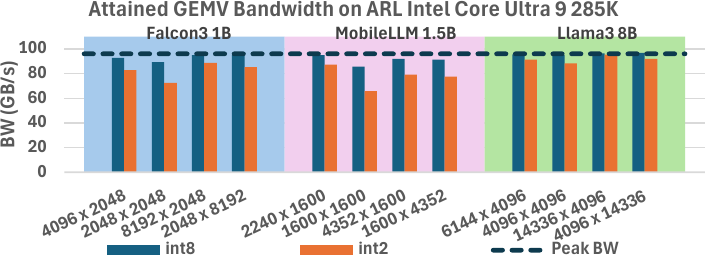}
\caption{Attained GEMV bandwidth on ARL for various matrix shapes and precisions (int8/int2).}
\label{fig:arl_micro}
\end{figure}

\begin{figure}[t]
\centering
\includegraphics[width=1.0\columnwidth]{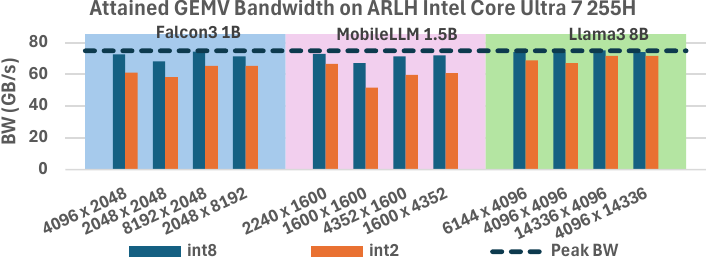}
\caption{Attained GEMV bandwidth on ARLH for various matrix shapes and precisions (int8/int2).}
\label{fig:arlh_micro}
\end{figure}

\begin{figure}[t]
\centering
\includegraphics[width=1.0\columnwidth]{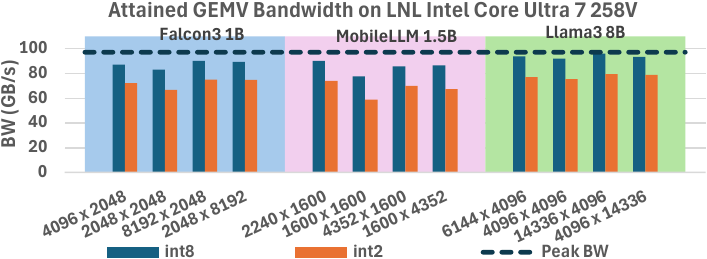}
\caption{Attained GEMV bandwidth on LNL for various matrix shapes and precisions (int8/int2).}
\label{fig:lnl_micro}
\end{figure}

Here we assess the efficacy of the ultra-low-bit GEMV kernels on the CPU platforms. We extracted the weight-matrix shapes from three models: Falcon3-1B~\cite{falcon} (blue plot areas), MobileLLM-1.5B~\cite{mobilellm} (magenta plot areas) and Llama3-8B model~\cite{llama8B} (green plot areas). These matrix shapes ($M\times K$) are shown in the x-axis of Figures~\ref{fig:arl_micro},~\ref{fig:arlh_micro} and~\ref{fig:lnl_micro}. The y-axis corresponds to the attained bandwidth (in GB/s) for the corresponding matrix-vector multiplication routines, and we show for each matrix shape two bars: i) With blue bar we depict the kernels with 8-bit (int8) weights, and ii) with orange bars we illustrate the kernels with 2-bit (int2) weights.

In Figure~\ref{fig:arl_micro} we illustrate the GEMV micro-benchmark results on ARL. As a baseline we consider the int8 GEMV performance where there is no up-convert overhead, and the GEMV microkernel consists of vector loads and VNNI-int8 FMAs. The int2 GEMV kernels are within 10--20\% of the int8 roofline and achieve bandwidth close to the machine's peak. For larger matrix sizes (right side of the plot), where the OpenMP/barrier overheads are less significant, the int2 GEMV achieves bandwidth within 2--5\% of the int8 GEMV. For the smallest matrix shape $1600\times 1600$, given the peak bandwidth 98 GB/s, the ideal int2 microkernel time is 6 microseconds, while we measure 8.9 microseconds. However, OpenMP parallel region fork/join and barrier overheads account for 1 microsecond as measured via benchmarking on ARL. In contrast, the int8 GEMV kernels move $4\times$ more data than the int2 ones, making the relative OpenMP overhead less important.

On ARL with total read bandwidth 98 GB/s and 24 cores, each core can draw 4 GB/s assuming equal bandwidth distribution. The operating frequencies of 5.4 GHz for $P$ cores and 4.5 GHz for $E$ cores (measured under full load) translate the per-core 4 GB/s to $\beta_P = 0.8$ bytes/cycle per $P$ core and $\beta_E = 0.95$ bytes/cycle per $E$ core (see dark magenta and light magenta vertical bars in Figure~\ref{fig:roofline}). According to the roofline, both $P$ and $E$ cores should achieve peak bandwidth (we reside in the slanted parts of the model), which aligns with our GEMV benchmark results.

Figure~\ref{fig:arlh_micro} shows GEMV results on ARLH. The conclusions are similar to ARL: int2 GEMV kernels are within 10--20\% of the int8 roofline and achieve near-peak bandwidth. With 75 GB/s total bandwidth and 14 cores, each core can draw $\sim$5.36 GB/s assuming equal bandwidth distribution. The operating frequencies of 4 GHz ($P$ cores) and 3.6 GHz ($E$ cores) yield $\beta_P = 1.44$ and $\beta_E = 1.6$ bytes/cycle (see dark and light blue vertical bars in Figure~\ref{fig:roofline}). Both $P$ and $E$ cores are modeled to hit peak bandwidth as confirmed by the benchmarks.

Finally, Figure~\ref{fig:lnl_micro} shows the GEMV microbenchmarks on LNL. We observe that the 2-bit microkernel delivers performance close to the peak (within 20--25\% for the larger matrix shapes). On LNL, with a total read bandwidth of 97 GB/s and 8 cores, the per-core bandwidth is approximately 12.1 GB/s. The measured operating frequencies of 4.1 GHz for $P$ cores and 3.7 GHz for $E$ cores translate to $\beta_P = 3.18$ bytes/cycle and $\beta_E = 3.52$ bytes/cycle (see dark red and light red vertical bars in Figure~\ref{fig:roofline}). We conclude that both $P$ and $E$ cores are able to reach close to peak read bandwidth, which is confirmed by the GEMV results, although we now operate much closer to the ``knee'' of the roofline for the E-cores.

\subsection{GEMV/GEMM microbenchmarks on Intel Xe2 GPUs}
\label{subsec:gpumicros}
\begin{figure}[t]
\centering
\includegraphics[width=1.0\columnwidth]{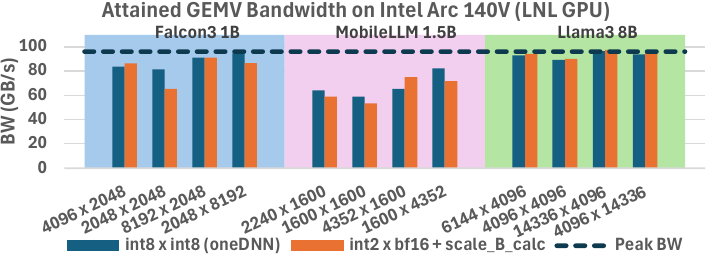}
\caption{Attained GEMV bandwidth on LNL-GPU for various matrix shapes and precisions (weights int2/int8).}
\label{fig:gemv_lnl_micro}
\end{figure}
Here we present analogous GEMV microbenchmarks for our Xe2 GPU systems. In addition to the memory bound GEMV microbenchmarks ($N=1$), we also present results for GEMM shapes with larger $N$ dimension that lead to compute bound kernels. The Xe2 GPUs offer high compute throughput via the int2$\times$int8$\rightarrow$int32  DPAS instructions (same as the int8$\times$int8$\rightarrow$int32 throughput, e.g.\ $\approx50$ TFLOPS for the integrated Intel Arc 140V and $\approx300$ TFLOPS for Battlemage B70), thus they are able to accelerate the compute bound/prefill phases of the int2 inference vs BF16 inference. Therefore, we also assess the efficacy of our mixed precision int2$\times$BF16$\rightarrow$BF16 Xe2 GPU kernels for compute-bound GEMMs.

In Figure~\ref{fig:gemv_lnl_micro} we illustrate the GEMV micro-benchmark results on the integrated Intel Arc 140V (LNL-GPU). As a baseline (blue bars) we consider the int8$\times$int8 GEMV performance from oneDNN where essentially there are no input/output quantization/dequantization overheads. The int2\ $\times$\ BF16$\rightarrow$BF16 GEMV kernels (including the calculation of the $B$ scales) are within 10\% of the int8 roofline and achieve bandwidth close to the machine's peak. For larger matrix sizes, where the kernel launch overheads are less significant, the int2 GEMV achieves the same bandwidth as the int8 GEMV. For the smallest matrix shape $1600\times 1600$, given the peak read bandwidth (97 GB/s), the ideal pure int2 microkernel time is 6 microseconds. The kernel launch overhead is on the order of 2 microseconds, thus we effectively observe $\sim$53 GB/s bandwidth. The sustained bandwidth by the int2 Xe2 LNL-GPU kernels is comparable and slightly higher than the one achieved by the LNL-CPU cores (see Figure~\ref{fig:lnl_micro}). This is expected since both CPU and Xe2 GPU have access to the same DDR5 memory, and the Xe2 kernels with the native int2$\times$int8 DPAS instructions are avoiding the up-convert overheads that the CPU kernels have. For all the GEMV cases, we found experimentally that fusing the $B$ scale calculation within the kernel (i.e.\ avoiding separate kernel launches at the expense of redundant computation) is beneficial, especially since the $B$ tensor is just an activation vector. These results indicate that our \emph{fused} int2\ $\times$\ BF16$\rightarrow$BF16 Xe2 kernels described in Section~\ref{subsec:gpu_kernels} are effectively hiding the quantization/dequantization overheads.
\begin{figure}[t]
\centering
\includegraphics[width=1.0\columnwidth]{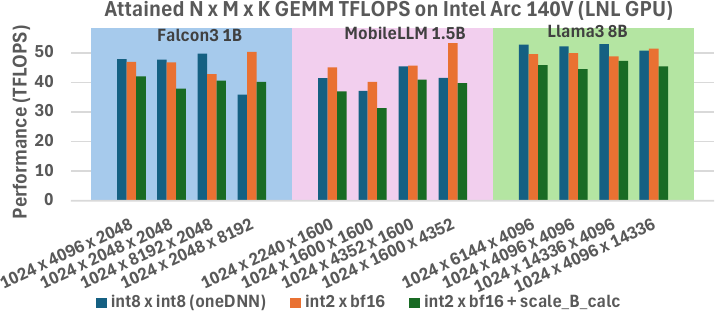}
\caption{Attained GEMM TFLOPS on LNL-GPU for various matrix shapes and precisions (weights int2/int8).}
\label{fig:gemm_lnl_micro}
\end{figure}

Figure~\ref{fig:gemm_lnl_micro} shows the attained GEMM performance ($N=1024$) for the same weight matrices as the ones in the previous microbenchmarks. These GEMMs have substantially higher weight re-use compared to the GEMV cases and are compute bound. To shed light on the performance we illustrate the TFLOPS for three configurations: i) a pure int8$\times$int8 GEMM from oneDNN without any quantization/dequantization overheads that should serve as an attainable roofline (blue bars), ii) the fused int2\ $\times$\ BF16$\rightarrow$BF16 Xe2 GEMM kernel only (illustrated in Figure~\ref{fig:gpu_kernel}, i.e.\ without the kernel/pass for calculating the $B$ scales) depicted with orange bars, and iii) the full-fledged mixed precision int2 GEMM including the overhead of calculating the $B$ scales via a separate kernel (green bars). First, we see that the fused mixed precision int2 GPU GEMM kernels achieve performance close to the pure int8$\times$int8 GEMM (note that the int2 $\times$ int8 DPAS instruction has the same throughput as the int8 $\times$ int8 DPAS). This result highlights the efficacy of our fused implementation that essentially hides all the online quantization and dequantization computations. For these \emph{compute-bound} GEMMs where the input activations are \emph{multiple} vectors, we found experimentally that the best performance is obtained by performing the input $B$ scale calculation as a \emph{separate kernel} before the GEMM (instead of fusing this $B$ scale calculation in the GEMM kernel as in the GEMV cases). This overhead is in the range of 5--15\% and for the cases with larger $M$ it diminishes to 3\%. Finally comparing the \emph{full} mixed precision kernel performance (green bars) with the performance of the pure int8 kernels from oneDNN serving as an attainable roofline (blue bars), we observe that the int2\ $\times$\ BF16$\rightarrow$BF16 Xe2 GEMMs are within 5--15\% of the pure int8 oneDNN GEMMs. For the larger tested matrices, we achieve on LNL-GPU up to 47 TFLOPS with the full-fledged int2 GEMM (whereas the max achieved is 53 TFLOPS by the int8 GEMM, being close to $\approx$ 90\% of the attainable peak). For comparison, LNL-CPU cores have a peak aggregate int8 FMA throughput of $\approx 1.7$ TFLOPS, therefore using the int2 GEMM on LNL-GPU Xe2 cores enables substantial benefits for compute bound cases (e.g.\ prefill stage, larger batch size).

\begin{figure}[t]
\centering
\includegraphics[width=1.0\columnwidth]{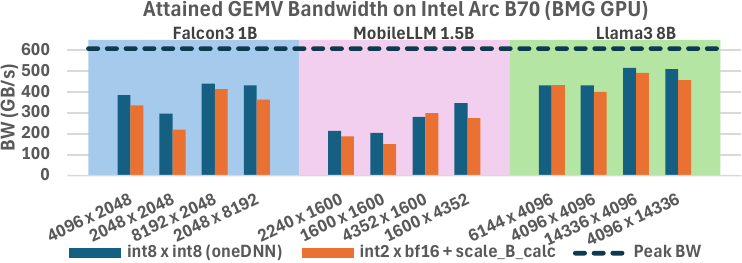}
\caption{Attained GEMV bandwidth on BMG-GPU for various matrix shapes and precisions (weights int2/int8).}
\label{fig:gemv_bmg_micro}
\end{figure}

\begin{figure}[t]
\centering
\includegraphics[width=1.0\columnwidth]{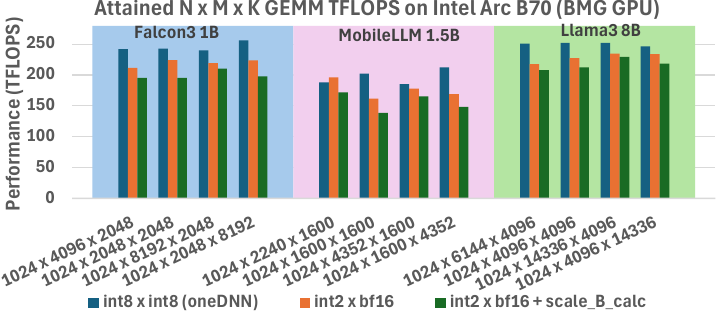}
\caption{Attained GEMM TFLOPS on BMG-GPU for various matrix shapes and precisions (weights int2/int8).}
\label{fig:gemm_bmg_micro}
\end{figure}

Figures~\ref{fig:gemv_bmg_micro} and~\ref{fig:gemm_bmg_micro} exhibit the analogous GEMV and GEMM results on the BMG discrete GPU, and the conclusions are similar to the LNL-GPU case: for both GEMV and GEMM cases the mixed precision int2\ $\times$\ BF16$\rightarrow$BF16 Xe2 GEMM performs (efficiency-wise) close to the pure int8$\times$int8 GEMM from oneDNN (representing an attainable roofline). For larger matrices, the ultra-low-bit kernels are within 5--10\% of the roofline. For the GEMV cases in Llama3-8B model, BMG achieves bandwidth up to 500 GB/s and for the GEMM cases ($N=1024$) we achieve up to 230 TFLOPS. Due to the increased bandwidth of BMG ($\sim$600 GB/s vs $\sim$97 GB/s on LNL-GPU) the kernel launch overheads are emphasized for the cases with small matrices (e.g.\ MobileLLM-1.5B model).

\subsection{End-to-end inference results on CPU}
\label{subsec:cpu_e2e}
We integrated our multi-threaded 2-bit GEMMs into the optimized PyTorch-TPP framework, which delivers state-of-the-art inference performance on CPUs~\cite{georganas2024harnessing}. First, we focus on three LLM models: Falcon3-1B~\cite{falcon}, MobileLLM-1.5B~\cite{mobilellm}, and Llama3-8B~\cite{llama8B}. The MobileLLM family, released by Meta as a companion to the recent ParetoQ work, is truly quantized to 2-bit using QAT techniques, achieving state-of-the-art accuracy for the respective model sizes~\cite{liu2025paretoq}. The same QAT techniques have been applied to Llama3 models, producing high-accuracy 2-bit variants, although the weights have not been released. Therefore, for performance evaluation, we use dummy 2-bit weights for Llama3-8B. Since bitnet.cpp has limited model coverage, we include Falcon3-1B (supported by bitnet.cpp) as the smallest model and also use dummy 2-bit weights for performance runs. All experiments generate 128 output tokens with batch size 1. For bitnet.cpp, we test various core configurations and report the best result. Additionally, since PyTorch-TPP supports MXFP4 (4-bit) Weight-Only Quantization (WOQ), we benchmark this variant as well~\cite{mlspecqd}.

\begin{figure}[t]
\centering
\includegraphics[width=1.0\columnwidth]{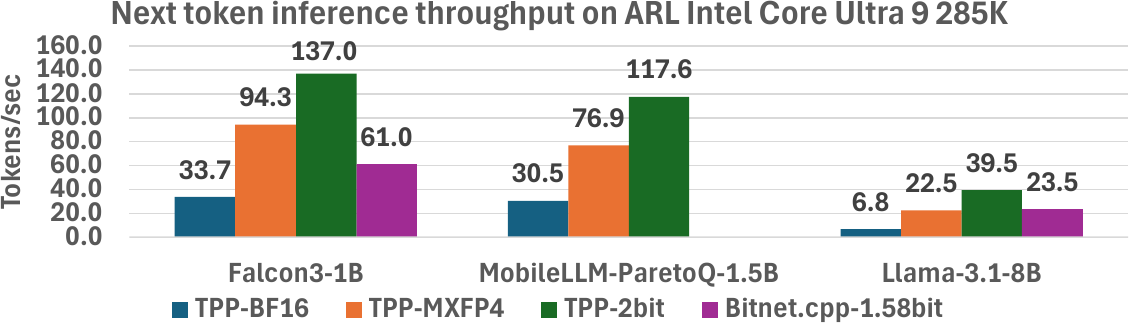}
\caption{End-to-end inference on ARL. For PyTorch-TPP inference: (i) Blue corresponds to BF16 weights. (ii) Orange corresponds to 4-bit weights. (iii) Green corresponds to 2-bit weights. Magenta shows the performance of \emph{ternary} bitnet.cpp.}
\label{fig:e2e_arl}
\end{figure}

Figure~\ref{fig:e2e_arl} illustrates the end-to-end inference results on ARL. For the inference with PyTorch-TPP we have three variants: (i) Dark blue bars correspond to BF16 weights. (ii) Orange bars correspond to MXFP4 weights. (iii) Green bars correspond to 2-bit weights. With magenta bars we represent the performance of the bitnet.cpp framework. The measured performance is in tokens/second (y-axis). First, we observe that the 4-bit inference with PyTorch-TPP is faster than bitnet.cpp on the 1B parameter model and on par with bitnet.cpp for the 8B model, even though bitnet.cpp uses 2-bit weights (and as such should be substantially faster than the 4-bit inference). These results suggest that bitnet.cpp is far from optimal. In our case, we observe with our optimized 2-bit inference substantial speedups over the 16-bit baseline, more precisely 4.1$\times$, 3.9$\times$ and 5.8$\times$ for the respective 1B, 1.5B and 8B models.

To understand these results, we derive simple performance models. Given the end-to-end inference execution we know that only the GEMV portions can be accelerated with our new kernels. Assuming that in the 16-bit inference the GEMV kernel represents a fraction $\alpha$ of the total time, and by applying the reasoning of Amdahl's law, the maximum attainable speedup $s$ over 16-bit inference by using datatype $x$ times smaller is:
\begin{equation}
    s = \frac{1}{1-\alpha+\frac{\alpha}{x}}
\end{equation}
For the 1B model, in the 16-bit run on ARL we get $\alpha=0.87$ (i.e.\ 87\% of the time is spent in the GEMVs), and as a result the maximum attainable speedup is 4.3$\times$ for the 2-bit inference ($x=8$). Under this lens, we indeed see that our obtained results are close to the ideal ones (we observed 4.1$\times$). At the other extreme (8B model), in the 16-bit run we get on ARL $\alpha=0.96$ (i.e.\ 96\% of the time is in GEMVs), thus the maximum attainable speedup is 6.1$\times$ for the 2-bit inference ($x=8$). We see that our obtained results are close to the ideal ones (we see 5.8$\times$ speedup over the 16-bit baseline). When comparing our 2-bit inference with the 1.58-bit inference of bitnet.cpp we observe on ARL speedups of 2.2$\times$ for the 1B model and 1.7$\times$ for the 8B case, showing that our optimized 2-bit kernels along with the PyTorch-TPP runtime substantially improve the SOTA 2-bit inference on CPUs.

\begin{figure}[t]
\centering
\includegraphics[width=1.0\columnwidth]{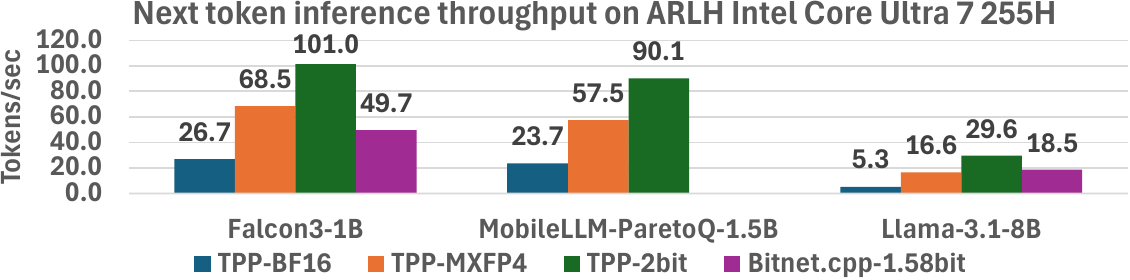}
\caption{End-to-end inference on ARLH.}
\label{fig:e2e_arlh}
\end{figure}
\begin{figure}[t]
\centering
\includegraphics[width=1.0\columnwidth]{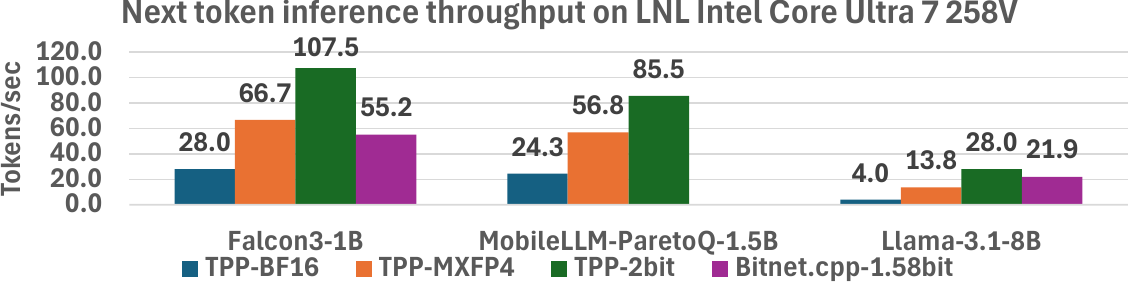}
\caption{End-to-end inference on LNL.}
\label{fig:e2e_lnl}
\end{figure}

Figure~\ref{fig:e2e_arlh} exhibits the end-to-end inference results on ARLH. The high-level picture is similar to the one described in the ARL case: we observe with our optimized 2-bit inference speedups over the 16-bit baseline, more precisely 3.8$\times$, 3.8$\times$ and 5.6$\times$ for the respective 1B, 1.5B and 8B models. When comparing our 2-bit inference with the 1.58-bit inference of bitnet.cpp we observe on ARLH speedups of 2$\times$ for the 1B model and 1.6$\times$ for the 8B case. Figure~\ref{fig:e2e_lnl} depicts the end-to-end inference results on LNL. We observe with our optimized 2-bit inference speedups over the 16-bit baseline, more precisely 3.8$\times$, 3.5$\times$ and 7$\times$ for the respective 1B, 1.5B and 8B models. When comparing our 2-bit inference with the 1.58-bit inference of bitnet.cpp we observe on LNL speedups of 1.9$\times$ for the 1B model and 1.3$\times$ for the 8B case.

\subsection{End-to-end inference results on Intel Xe2 GPUs}
\label{subsec:gpu_e2e}
\begin{figure}[t]
\centering
\includegraphics[width=1.0\columnwidth]{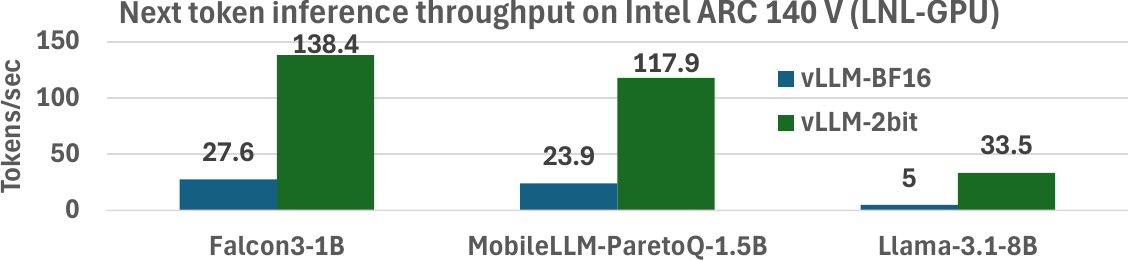}
\caption{End-to-end inference on LNL-GPU.}
\label{fig:e2e_lnl_gpu}
\end{figure}

\begin{figure}[t]
\centering
\includegraphics[width=1.0\columnwidth]{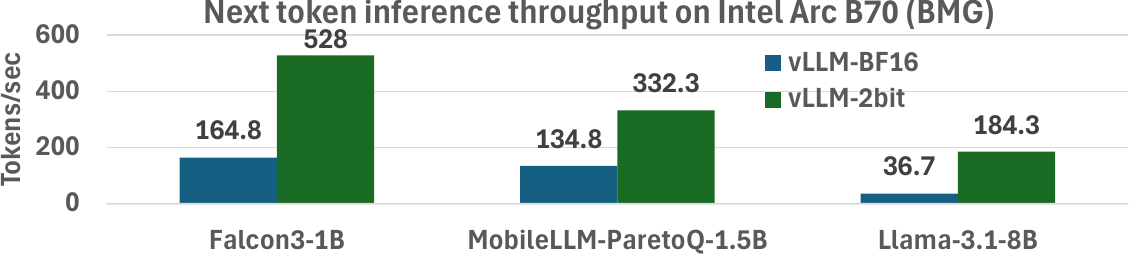}
\caption{End-to-end inference on BMG-GPU.}
\label{fig:e2e_bmg_gpu}
\end{figure}

For the evaluation of the end-to-end inference with the Intel Xe2 GPUs, we integrated our kernels into the vLLM~\cite{vllm} framework as a quantization plugin and evaluated the inference on the same models and setup described in Section~\ref{subsec:cpu_e2e}. We leveraged the XPUGraph runtime optimizations~\footnote{\url{https://docs.pytorch.org/docs/stable/generated/torch.xpu.XPUGraph.html}} to reduce kernel host overhead on Xe2 GPU devices. Figure~\ref{fig:e2e_lnl_gpu} exhibits the end-to-end inference on LNL-GPU with BF16 (blue) and 2-bit weights (green), and we observe that the 2-bit inference provides speedup up to 6.7$\times$ for the inference with the Llama3-8B model, achieving for this case 33.5 tokens/second. As expected, this throughput is in alignment with the obtained LNL-CPU result ($\sim$28 tokens/second) since both CPU and Xe2 GPU on the LNL platform have access to the same bandwidth. However, on the LNL platform one can benefit from leveraging the Xe2 GPU in compute-bound use-cases like inference with longer context, larger batch size and shorter output length, since the compute-bound GEMM performance gap of LNL-CPU and LNL-GPU is $\sim$30$\times$ as shown in Section~\ref{subsec:gpumicros}. We ran an experiment with the 2-bit Llama3-8B model, 2048 input tokens and 16 output tokens, and we observed that LNL-GPU is $\sim26\times$ faster than LNL-CPU considering end-to-end latency.

Figure~\ref{fig:e2e_bmg_gpu} illustrates the end-to-end inference on BMG for the same experimental setup. Inference with the largest model, Llama3-8B, is accelerated by 5$\times$ using our optimized 2-bit kernels (compared to the BF16 inference). For the case with the smaller MobileLLM-1.5B model, the observed speedup on BMG is 2.5$\times$. As shown in the magenta part of Figure~\ref{fig:gemv_bmg_micro}, these relatively small int2 GEMVs achieve 150--300 GB/s due to the emphasized kernel overheads. On the other hand, the corresponding BF16 GEMVs achieve $\sim$500 GB/s (less emphasized kernel overheads due to the larger duration), thus the end-to-end overall benefit is 2.5$\times$.

\subsection{End-to-end inference with production-quality ternary Bonsai models}
\label{subsec:bonsai_e2e}
\begin{figure}[t]
\centering
\includegraphics[width=1.0\columnwidth]{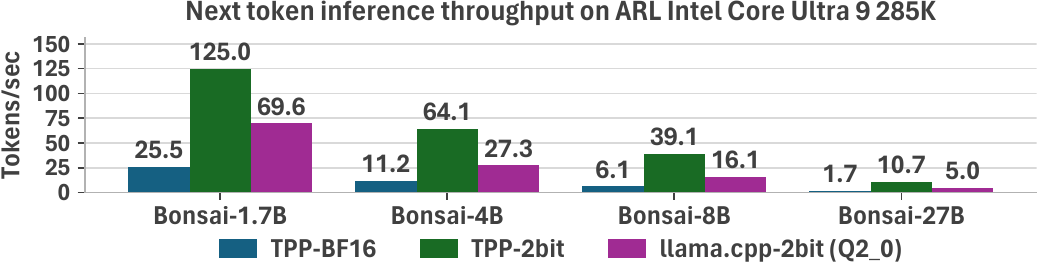}
\caption{End-to-end inference on ARL (ternary Bonsai).}
\label{fig:e2e_arl_bonsai}
\end{figure}
\begin{figure}[t]
\centering
\includegraphics[width=1.0\columnwidth]{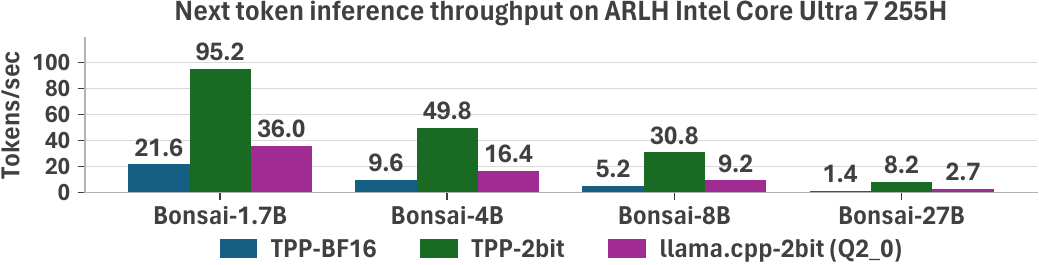}
\caption{End-to-end inference on ARLH (ternary Bonsai).}
\label{fig:e2e_arlh_bonsai}
\end{figure}
\begin{figure}[t]
\centering
\includegraphics[width=1.0\columnwidth]{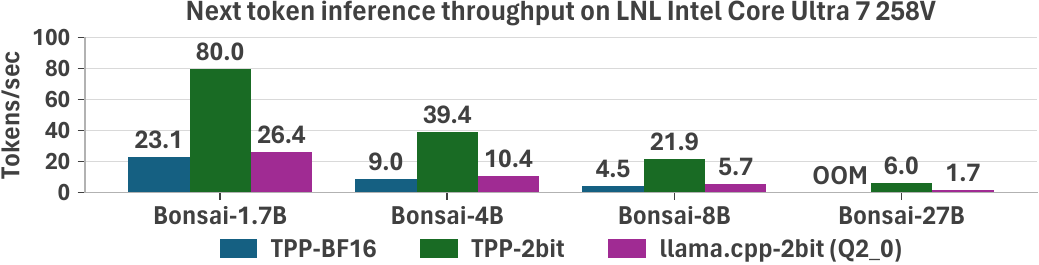}
\caption{End-to-end inference on LNL (ternary Bonsai).}
\label{fig:e2e_lnl_bonsai}
\end{figure}
\begin{figure}[t]
\centering
\includegraphics[width=1.0\columnwidth]{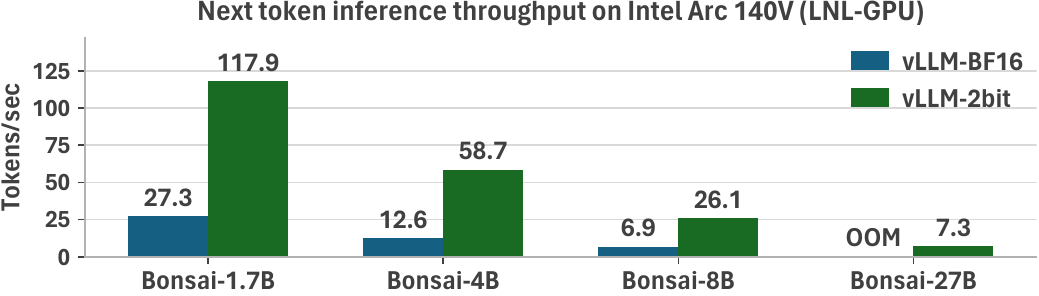}
\caption{End-to-end inference on LNL-GPU (ternary Bonsai).}
\label{fig:e2e_lnlgpu_bonsai}
\end{figure}
\begin{figure}[!h]
\centering
\includegraphics[width=1.0\columnwidth]{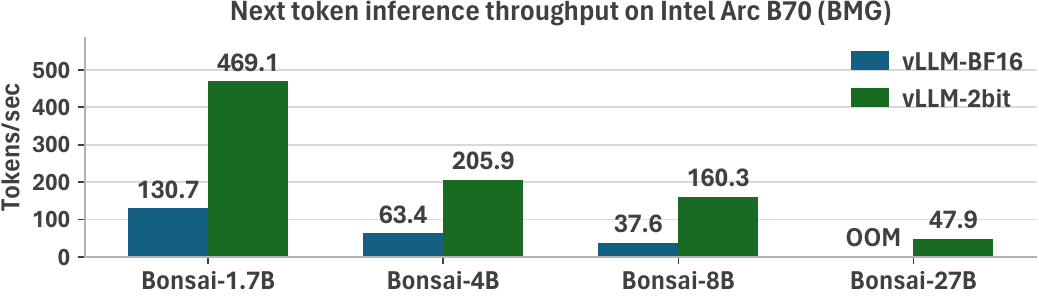}
\caption{End-to-end inference on BMG-GPU (ternary Bonsai).}
\label{fig:e2e_b70_bonsai}
\end{figure}
In this section we show end-to-end inference results with the state-of-the-art Bonsai ternary LLM models~\cite{prismml2026bonsai}. These production-quality ternary models are competitive with their full-precision counterparts (e.g.\ the 27B model is within $\sim$5\% accuracy of the 16-bit Qwen3.6-27B), while showing almost a 10$\times$ increase in \emph{intelligence density} (i.e.\ intelligence per unit of storage) over the 16-bit counterparts~\cite{prismml2026bonsai}. With respect to the quantization recipe, the Bonsai weights are quantized with a group size of 128 (i.e.\ each group of 128 weights along the contraction dimension $K$ shares a common scale) and the corresponding weight scales are stored in 16-bit precision.

We experimented with four ternary Bonsai models: 1.7B, 4B, 8B and 27B. Figures~\ref{fig:e2e_arl_bonsai}, \ref{fig:e2e_arlh_bonsai} and \ref{fig:e2e_lnl_bonsai} show the end-to-end inference performance on the ARL, ARLH and LNL CPU platforms respectively. The blue bars correspond to the reference 16-bit (BF16) implementation, while the green bars correspond to the ternary (2-bit) Bonsai models with our optimized 2-bit inference kernels and the PyTorch-TPP runtime. We observe that, depending on the model size and the platform, we obtain speedups in the range of 3.5--6.4$\times$ over the 16-bit inference. We note that on the LNL platform, which has 32 GB of memory available, the largest 27B model cannot run with the full BF16 precision weights since that would require $\sim$54 GB of memory. Nevertheless, the ternary 27B model requires only $\sim$7 GB of memory and is able to run on the memory-constrained LNL platform, achieving a decoding throughput of 6 tokens/second.

The Bonsai model release is accompanied by a fork of the \texttt{llama.cpp}~\cite{llamacpp} runtime that is able to serve the ternary models on x86 CPUs with a 2-bit storage datatype (referred to as \texttt{Q2\_0} in \texttt{llama.cpp}). In Figures~\ref{fig:e2e_arl_bonsai}, \ref{fig:e2e_arlh_bonsai} and \ref{fig:e2e_lnl_bonsai} we illustrate with magenta bars the corresponding achieved decoding throughputs. For the \texttt{llama.cpp} runs we experimented with three core configurations pertaining to performance ($P$) and efficiency ($E$) cores, namely $P$-only, $E$-only and $P$+$E$, and we report the best performance: the $P$+$E$ configuration is optimal on ARL and LNL, whereas on ARLH the $E$ cores degrade the throughput and the $P$-only configuration performs best. We conclude that the corresponding 2-bit kernels in \texttt{llama.cpp} are far from optimal: on ARL our 2-bit inference is 1.8--2.4$\times$ faster than the 2-bit inference with \texttt{llama.cpp}, on ARLH we observe speedups in the range of 2.7--3.3$\times$, and on LNL the performance benefit in the decoding stage is 3--3.9$\times$. These results highlight the significance of our bottom-up approach that yields 2-bit kernels running close to the roofline.

Figures~\ref{fig:e2e_lnlgpu_bonsai} and~\ref{fig:e2e_b70_bonsai} illustrate the end-to-end inference on the LNL-GPU and the BMG Xe2 platforms. Note that \texttt{llama.cpp} does \emph{not} include optimized 2-bit kernels for Xe2 architectures, thus here we simply compare our optimized 2-bit Xe2 inference in vLLM with the 16-bit baseline. On LNL-GPU we observe speedups of 3.8--4.7$\times$ over the 16-bit baseline. We note again that the 27B model runs out of memory with the 16-bit precision, whereas the 2-bit inference achieves 7.3 tokens/second; this performance is in alignment with the LNL-CPU execution (6 tokens/second) since both the CPU and the integrated GPU have access to the same memory bandwidth. Last, in Figure~\ref{fig:e2e_b70_bonsai} we see that on the discrete BMG GPU we achieve speedups in the range of 3.3--4.3$\times$ over the 16-bit execution (again the 27B 16-bit model does not run on this platform, which has 32 GB of memory). By comparing the int2 decode throughput achieved on BMG with the one on LNL-GPU, we observe that for the largest 27B model BMG is $\sim$6.5$\times$ faster than LNL-GPU, which is in alignment with the bandwidth gap of these two platforms. It is worth mentioning that even for such a production-quality, highly capable 27B parameter model we achieve throughput on the order of $\sim$48 tokens/second, making LLM inference practical for interactive use cases on a client-class discrete GPU.

\subsection{CPU comparison with 2-bit inference on NVIDIA GPUs}
\label{subsec:a100_comparison}
Bitnet.cpp includes CUDA-optimized 2-bit kernels. To make an apples-to-apples comparison with our CPU results, we benchmarked our 2-bit inference pipeline on a same-size 2 billion parameter model\footnote{\url{https://huggingface.co/andrijdavid/Llama3-2B-Base}}, with 32 input tokens and 128 output tokens such that we exercise the decoding phase of inference. On our NVIDIA A100 GPU and with the bitnet-b1.58 2 billion model we were able to obtain 300 tokens/second (in alignment with the reported performance on the bitnet repo), whereas we achieve on the aforementioned same-sized 2B model 110 tokens/s on ARL, 82 tokens/s on ARLH, and 88 tokens/s on LNL. Thus, this level of performance on the CPUs is within 2.7$\times$--3.7$\times$ of the A100 GPU which has 16$\times$--21$\times$ more bandwidth than our CPUs (A100 has 1.6 TB/s bandwidth vs 75--98 GB/s CPU bandwidth). To further shed light on these results, we assessed the efficacy of the 2-bit bitnet.cpp GEMV kernels on A100 compared to the corresponding 16-bit ones (these GEMV results are also provided in the bitnet.cpp repo). On the A100 GPU, for the smaller matrix shapes (e.g.\ 2560$\times$2560, 3840$\times$2560, 3200$\times$3200) the 2-bit GEMV speedup over the 16-bit GEMV is only 1.6$\times$--2.1$\times$ whereas for the larger shapes the speedup goes up to 3.9$\times$ (see Figure~\ref{fig:gemv_comparison}, bf16 GEMV performance on A100 with dark magenta bars and int2 GEMV on A100 with light magenta bars). Considering that the majority of matrices in such LLMs (i.e.\ 2B) are relatively small, we expect the upside of the 2-bit GEMVs to be limited on A100. For comparison, in similarly-sized matrices we achieved close-to-ideal bandwidth with our 2-bit GEMV kernels on CPUs, meaning that they are $\approx7\times$ faster than the corresponding 16-bit GEMVs, and as a result we materialized substantial speedup in the end-to-end 2-bit inference. We also experimented with the 2-bit GPU inference from bitnet.cpp using a recent RTX PRO 6000 Blackwell Server Edition (henceforth called B6000). On B6000 we observed for the same setup 405 tokens/second, and consequently our CPUs are within 3.7--4.9$\times$ of B6000, despite B6000 offering 1.8 TB/s bandwidth vs 75--98 GB/s CPU bandwidth (i.e.\ 18--24$\times$ gap).

\subsection{BMG comparison with 2-bit inference on NVIDIA GPUs}
\begin{figure*}[t]
\centering
\includegraphics[width=1.0\textwidth]{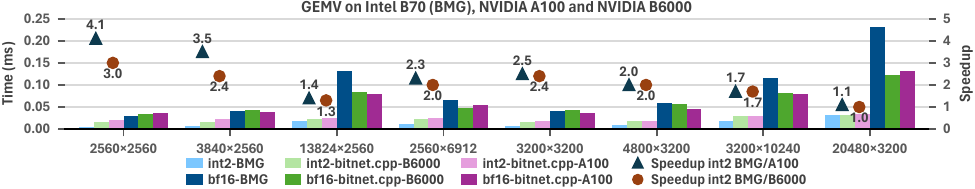}
\caption{GEMV time on NVIDIA-A100, NVIDIA-B6000, Intel-B70 (BMG) for various matrix shapes $M\times K$ shown on the x-axis and precisions. The right axis reports the int2 speedup of BMG over A100 (triangles) and over B6000 (circles).}
\label{fig:gemv_comparison}
\end{figure*}
Finally, we compare the 2-bit end-to-end inference performance on B70 (BMG) with the NVIDIA A100 and B6000 2-bit inference from bitnet.cpp on a same-sized 2B model (see Table~\ref{tab:gpuperf}, same setup as the one in Section~\ref{subsec:a100_comparison}). On BMG we obtained 542 tokens/second, being 1.8$\times$ \emph{faster} than the 2-bit inference on A100 and 1.34$\times$ \emph{faster} than the 2-bit inference on B6000, despite the fact that BMG has 2.7--3$\times$ lower bandwidth than A100 and B6000. To understand this behavior, we illustrate in Figure~\ref{fig:gemv_comparison} the GEMV times for various shapes (matrix dimensions extracted from bitnet.cpp) and precisions on BMG, A100 and B6000. First, by comparing the BF16 GEMV times (dark blue bars for BMG, dark magenta bars for A100, dark green bars for B6000) we observe that A100 and B6000 are for larger matrices up to 2$\times$ faster than BMG. However, when comparing 2-bit GEMV times (light blue bars for BMG, light magenta bars for A100, light green bars for B6000) the landscape changes: BMG is up to 4.1$\times$ \emph{faster} than A100 (see triangle speedup points) and up to 3$\times$ \emph{faster} than B6000 (see circle speedup points) despite having 2.7$\times$ and 3$\times$ less bandwidth respectively. The efficiency of our 2-bit Xe2 GEMV kernels (e.g.\ see Figure~\ref{fig:gemv_bmg_micro}) is higher than that of the 2-bit A100 and B6000 GEMV kernels (which show only limited speedup of 1.6$\times$--2.7$\times$ over the BF16 GEMV for smaller matrix shapes). As a result, the 2-bit end-to-end inference on BMG ends up being faster than the 2-bit inference on A100 and B6000 for a 2B parameter model.

\subsection{Power efficiency of 2-bit inference on GPUs}
\begin{table}[t]
  \begin{center}
    \footnotesize
    \begin{tabular}{c| c|c |c}
      \textbf{GPU} & \textbf{Tokens/s} &\textbf{Power (W)} &\textbf{Tokens/J}\\\hline \hline
      Intel 140V (LNL-GPU) & 106.7 & 20 & 5.34\\ \hline
     Intel B70 (BMG) & 542 & 190 & 2.85\\ \hline
    NVIDIA A100 & 300 & 155 & 1.94\\ \hline
   NVIDIA B6000 & 405 & 230 & 1.76\\ \hline
    \end{tabular}
  \end{center}
\caption{GPU 2-bit inference on a 2 billion parameter model}
\label{tab:gpuperf}
\end{table}
We also investigated the power efficiency of the int2 GPU inference (Table~\ref{tab:gpuperf}). Regarding the methodology, we launched long-running experiments with a 2 billion parameter model and 8192 output tokens, and measured the reported power consumption (via the nvtop Linux tool) at steady state. For the LNL-GPU we report the total package power (i.e.\ including the CPU power) whereas for the rest of the cases we report only the GPU power. By using tokens/J as the figure of merit, we see that the integrated LNL-GPU is the most power-efficient, outperforming the high-end A100 and B6000 GPUs by 2.75$\times$ and 3$\times$ respectively. The discrete Intel B70 GPU achieves 2.85 tokens/J and outperforms A100 and B6000 by 1.47$\times$ and 1.6$\times$ in power efficiency. Thus we conclude that our 2-bit inference on an edge integrated GPU (LNL-GPU) outperforms high-end GPUs by up to 3$\times$ in tokens/J while being within 3.8$\times$ in token throughput (despite having 18$\times$ less bandwidth).

\section{Related Work}
\label{sec:related}
\subsection{LLM Inference and Ultra-Low-Bit Quantization}
In recent years, several methods have been developed to reduce LLM inference costs~\cite{treviso2023efficient}, minimize latency and increase efficiency, including sparsification/pruning~\cite{frantar2023sparsegpt}, knowledge distillation~\cite{hinton2015distilling,taori2023stanford} and quantization~\cite{lin2024awq, chee2023quip, liu2025paretoq}. Among these, quantization reduces the precision of model weights (e.g.\ from 16 bits to 4/2/1 bits), thereby lowering memory requirements and potentially accelerating inference, especially when the target hardware supports specialized instructions or when the pipeline operates in a bandwidth-bound regime~\cite{lang2024comprehensive}. Quantization methods for LLMs fall into two categories: Quantization-Aware Training (QAT) and Post-Training Quantization (PTQ)~\cite{hasan2024optimizing}. QAT incorporates reduced-precision weights directly into the model's pretraining/fine-tuning. By accounting for quantization effects during training, QAT enables the model to adjust its parameters with awareness of limited precision, resulting in better accuracy. In contrast, PTQ applies quantization to a pretrained model. Although QAT achieves higher accuracy, it is more computationally intensive, requiring access to representative training datasets~\cite{liu2025paretoq}.

As LLMs continue to scale in size and training data, increasing attention has been drawn to scaling laws that balance model/dataset size to optimize performance and computational efficiency~\cite{hoffmann2022training, kumar2024scaling,dettmers2023case}. In parallel, the field is moving towards ultra-low-bit LLMs, motivated by the gains in memory and compute efficiency~\cite{liu2023binary, 1bit, liu2025paretoq}. This trend calls for a re-evaluation of scaling laws to consider the impact of quantization on model performance. Recent work shows that 1.58-bit (ternary) and 2-bit quantization outperform both 4-bit and binary quantization in the size-accuracy trade-off~\cite{liu2025paretoq}. On this ultra-low-bit front, recent advances in both PTQ~\cite{wang2026cat,zhao2026twla} and QAT~\cite{liu2025paretoq,prismml2026bonsai,1bit,kaushal2024spectra,bitcpmcann8b,huang2026tequila} methods yield quantized models that come within 5--10\% of the perplexity and end-task performance of their full-precision counterparts. These techniques are essential and complementary to our work: they provide high-quality sub-4-bit models whereas our work with ultra-low-bit kernels enables efficient 2-bit runtime inference.

\subsection{Accelerating ultra-low-bit LLM Inference}
Despite the promising results of ultra-low-bit LLMs, practical inference with such models remains challenging. Recent work, such as bitnet.cpp~\cite{bitnet}, introduced an inference engine tailored for BitNet b1.58~\cite{1bit} and ternary LLMs. Bitnet.cpp includes a custom GEMM designed for efficient, lossless inference with sub-2-bit weights. While bitnet.cpp improves upon existing frameworks like \texttt{llama.cpp}~\cite{llamacpp} (up to 6$\times$ speedup thanks to the specialized GEMM kernel), its efficiency has not been assessed in terms of optimality. Our bottom-up approach in this work shows that close-to-optimal 2-bit microkernels can yield speedups of up to 2.2$\times$ over bitnet.cpp. Even the fork of \texttt{llama.cpp} that serves as the companion runtime for the Bonsai models~\cite{prismml2026bonsai} delivers poor performance on the CPU platforms we evaluated, underperforming our proposed end-to-end 2-bit inference by 1.8--3.9$\times$. Prior research has focused on lookup-table (LUT)-based mixed-precision GEMMs in deep learning~\cite{ganji2023deepgemm,blalock2021multiplying, tang2023lut, wei2407t}, especially for ultra-low-bit widths. In contrast, our work demonstrates that Fused Multiply-Add (FMA)-based approaches, paired with well-designed tensor-layouts and microkernels, achieve near-roofline performance and push the envelope of LLM inference. For GPUs, prior work applied intra-weight mixed precision quantization for weights with 2-bit and 4-bit groups in order to control the accuracy loss~\cite{li2024fast}. With recent advancements in QAT, 2-bit weight quantization is delivering robust accuracy, and the bitnet.cpp runtime includes such int2 GEMM kernels for CPUs and NVIDIA GPUs. To the best of our knowledge, the 2-bit kernels presented in this work are the fastest for x86 CPUs and Intel Xe2 GPUs, delivering close-to-peak performance, as illustrated by our benchmarking.

\section{Conclusions and Future Work}
\label{sec:conclusions}
In this work, we designed and implemented 2-bit GEMM kernels for modern x86 CPUs, achieving close to roofline efficiency on a variety of CPUs. We integrated these kernels into a SOTA LLM inference framework and presented end-to-end inference results with 2-bit models that outperform the SOTA runtime (bitnet.cpp) by up to 2.2$\times$, and deliver up to 7$\times$ speedup compared to 16-bit model inference. We also extended this work to Intel Xe2 GPUs where we implemented mixed precision, 2-bit GEMM kernels, and showed their performance to be close-to-optimal, accelerating BF16 inference by up to 6.7$\times$. On an Intel B70 Xe2 GPU we achieve a speedup of 1.8$\times$ over the 2-bit inference on NVIDIA A100 and 1.34$\times$ over the 2-bit inference on NVIDIA B6000 for a same-sized 2 billion-parameter model, despite A100/B6000 having 2.7/3$\times$ more bandwidth than Intel B70. Thus, our work pushes the envelope of LLM inference, and demonstrates that ultra-low-bit inference on CPUs and discrete client GPUs can approach or exceed high-end GPU-level performance. We plan to extend our work to ARM and NVIDIA/AMD platforms: our 2-bit kernels can be translated to AArch64 and SVE instructions on modern ARM CPUs, and the same fusion techniques are applicable for CUDA-based/ROCm-based 2-bit GEMM kernels.

\bibliographystyle{unsrt}
\bibliography{references}

\scriptsize
\noindent
\newline Optimization Notice: Software and workloads used in
performance tests may have been optimized for performance only on
Intel microprocessors.  Performance tests, such as SYSmark and
MobileMark, are measured using specific computer systems,
components, software, operations and functions.  Any change to any
of those factors may cause the results to vary.  You should
consult other information and performance tests to assist you in
fully evaluating your contemplated purchases, including the
performance of that product when combined with other products.
For more information go to \url{http://www.intel.com/performance}.

\noindent Intel, Xeon, and Intel Xeon Phi are trademarks of Intel Corporation in the U.S. and/or other countries.

\normalsize

\end{document}